\documentclass{article}

\usepackage{arxiv}

\usepackage[utf8]{inputenc} % allow utf-8 input
\usepackage[T1]{fontenc}    % use 8-bit T1 fonts
\usepackage{hyperref}       % hyperlinks
\usepackage{url}            % simple URL typesetting
\usepackage{booktabs}       % professional-quality tables
\usepackage{amsfonts}       % blackboard math symbols
\usepackage{nicefrac}       % compact symbols for 1/2, etc.
\usepackage{microtype}      % microtypography
\usepackage{lipsum}
\usepackage{graphicx}
\usepackage{float}
\usepackage{adjustbox}
\usepackage{multirow}
\graphicspath{ {./images/} }

\title{End-to-end Clinical Event Extraction from Chinese Electronic Health Record}

\author{
 Wei Feng \\
  Department of Medical Informatics\\ 
  School of Biomedical Engineering and Informatics\\
  Nanjing Medical University\\
   \And
 Ruochen Huang \\
  Department of Medical Informatics\\ 
  School of Biomedical Engineering and Informatics\\
  Nanjing Medical University\\
  \And
 Yun Yu \\
  Department of Medical Informatics\\ 
  School of Biomedical Engineering and Informatics\\
  Nanjing Medical University\\
  \And
 Huiting Sun \\
  Department of Medical Informatics\\ 
  School of Biomedical Engineering and Informatics\\
  Nanjing Medical University\\
  \And
 Yun Liu* \\
  Department of Information\\
  the First Affiliated Hospital \\
  Nanjing Medical University\\
  \texttt{liuyun@njmu.edu.cn} \\
  \texttt{*Correspondence Author} 
}

\begin{document}
\maketitle
\begin{abstract}
Event extraction is an important work of medical text processing. According to the complex characteristics of medical text annotation, we use the end-to-end event extraction model to enhance the output formatting information of events. Through pre training and fine-tuning, we can extract the attributes of the four dimensions of medical text: anatomical position, subject word, description word and occurrence state. On the test set, the accuracy rate was 0.4511, the recall rate was 0.3928, and the F1 value was 0.42. The method of this model is simple, and it has won the second place in the task of mining clinical discovery events (task2) in the Chinese electronic medical record of the seventh China health information processing Conference (chip2021).
\end{abstract}

% keywords can be removed

\section{Introduction}
Electronic health record (EHR) is composed of unstructured text and structured data. Therefore, information extraction is one of the key tasks in processing unstructured text in EHR. The task of EHR information extraction is mainly concentrated in the drug \cite{wei_study_2020, baer_can_2016, zheng_medication_2015} and disease fields \cite{sada_validation_2016, xu_extracting_2011, liu_use_2021}. 

Event extraction is an important task among information extraction tasks. General event extraction task is mostly based on the identification or classification of trigger words, event types, event elements, arguments, etc. \cite{zhan_survey_2019}. The event extraction tasks can play a key role in question answering, knowledge extraction, and knowledge map construction \cite{berant_modeling_2014}. Most of these tasks decomposed the event task into multiple sub-tasks, including extraction of trigger words, extraction of attributes, and merging of the sub modules. Such architecture requires high quality annotation of the original sentence, but it will become a time-consuming and laborious problem in medical treatment. In this task, we applied the end-to-end generation model to output the event extraction information into a structured and enhanced character string, and obtained the score of F1 0.42 in task 2 of the 2021 China health information processing Conference (chip2021)\footnote{http://cips-chip.org.cn}, ranking second.

\section{Related works}
Traditional event tasks mostly contained classifiers based on pattern recognition or machine learning methods, such as Monte Carlo Gibbs sampling \cite{finkel_incorporating_2005}, conditional random fields \cite{finkel_exploiting_2004}, support vector machines \cite{walsh_identifying_2017}, and so on. With the extensive application of deep learning, deep neural network model is also more and more applied to the task of event extraction, such as convolutional neural network \cite{chen_event_2015} and graph neural network \cite{liu_jointly_2018}. In medical text event extraction, traditional rule-based models \cite{tian_automated_2017, nath_natural_2016, lin_tepapa_2017} were mostly used, and the event extraction model based on deep learning method \cite{shi_multiple_2016} had also been popularized in recent years.

Most of these studies used the step-by-step event detection paradigm, that is, the detection of trigger words of events and the detection of arguments (event attributes). In this way, the two tasks were decomposed, and it was difficult to correlate them. Moreover, most of these models were too complex to do further task processing. Therefore, we combined the trigger word (core word) and argument (event attribute) of an event into one task, used the end-to-end paradigm as seq2seq \cite{sutskever_sequence_2014} to output both at the same time.

Seq2seq model contains encoder and decoder modules. The encoder module accepts the input sequence $W=w_1w_2w...w_n$ coded as an implicit tensor $H=h_1h_2...h_n$. The target sequence is then output through the Decoder module. On this basis, Google AI proposed a Text-to-Text Transfer Transformer (T5) \cite{raffel_exploring_2020} model. The T5 model combines the advantages of the Transformer architecture \cite{vaswani_attention_2017} to unify natural language processing tasks into end-to-end tasks. The T5 model using the Transformer architecture uses stacked multi-headed attention mechanisms, considers relative character location information, and fuses context semantics to achieve global information learning.

\section{Methods}
The objective of the CHIP2021 task2 task is to extract clinical discovery events from Chinese electronic medical records, all of which come from real medical data. For example, "\emph{..., the above symptoms occur repeatedly, without obvious incentives for each attack, they occur suddenly and last for several minutes...}" should extract event attributes such as "\emph{core name: symptoms, tendencies: yes, characteristics: recurrence, no incentives, sudden occurrence}". Each event of a dataset has four attributes (\emph{core name, trendency, role, and anatomy}). Roles and anatomies can be multiple entities in each core name.

\begin{figure}[H]
  \centering
  \includegraphics[width=7cm]{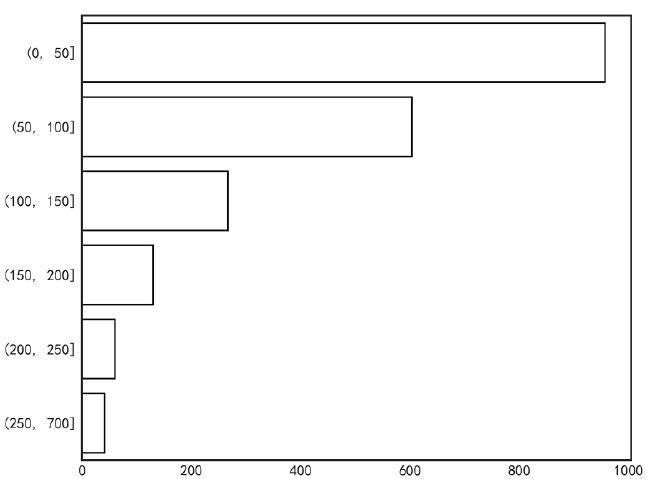}
  \caption{Overall distribution of medical record length.}\label{fig:len}
\end{figure}

The medical record data set of chip2021 task2 was used in this study. The data set has 2060 medical records, including 11129 events. The length distribution of medical records in the data set is shown in Figure \ref{fig:len}. The length of more than 95\% of sentences is within 200. 1854 training sets were randomly extracted, and the remaining 206 were used as verification sets. The training set contained 10135 events and the training set contained 994 events.

%  To find a model that fits the task, we analyzed the structure of the data and found that the percentage of custom events was lower than 12\% (Table \ref{tab:desc_ratio}). To further observe the custom words, for example, for the label data, the core word is pain, not in the original sentence, for the "outpatient department to our department with vascular headache"; "No systemic yellow dye", marked as "systemic" and so on. For core words, there are 11,129 core words with a low proportion of customizations, so the generated model is expected to fit this task better.

%  \begin{table}[H]
%   \caption{The proportion of events not in the original sentence.}
%    \centering
%    \begin{tabular}{lll}
%     \toprule
%     & Amount & Ratio  \\
%     \midrule 
% Core words & 1259   & 0.1131 \\
% Characters & 601    & 0.1178 \\
% Anotomy    & 438    & 0.1054 \\
% \bottomrule
% \end{tabular}
%    \label{tab:desc_ratio}
%  \end{table}

 \begin{figure}[H]
  \centering
  \includegraphics[width=7cm]{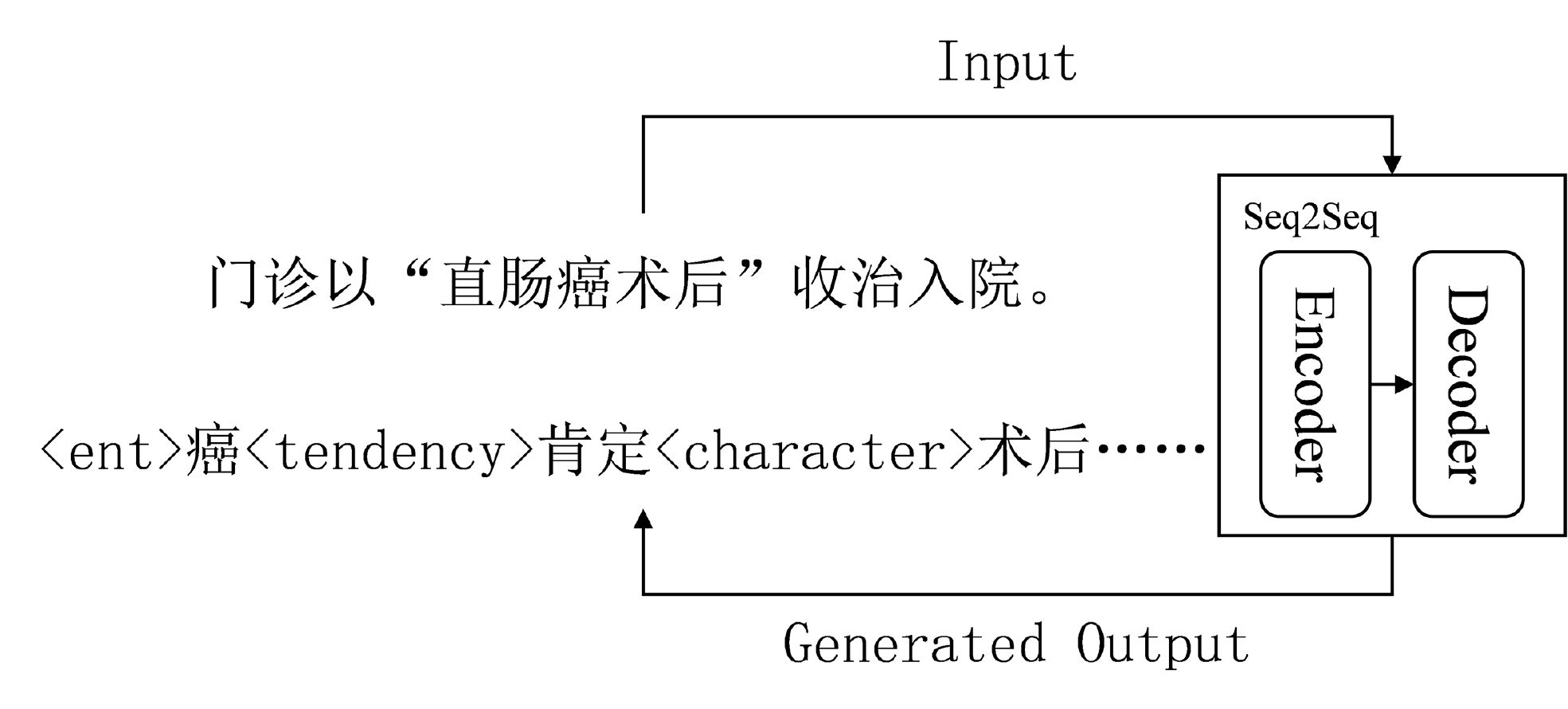}
  \caption{Structure of event text generation model.}\label{fig:process}
\end{figure}

Figure \ref{fig:process} describes the brief structure of our model. The input was a medical sentence from EHR. After trained through a seq2seq model, the output was our customized structured text. The input text waas formatted as $W=w_1w_2...w_n$ , where each $w_i$ represents the $i^{th}$ input word item, and $n$ represents the number of word items of the input sentence. The goal of the model was to extract all attributes of a event which was formatted as $A_j=a_j^1a_j^2...a_j^{m_j}$, where $j$ represents the $j^{th}$ attribute, $m_j$ represents the length of the attribute. 

\begin{table}[H]
  \caption{Input output examples for baseline and our models.}
  \begin{adjustbox}{max width=\textwidth}
   \begin{tabular}{lp{5cm}l}
    \toprule
    models   & input                                                                            & output                                                                                                                                                      \\ \midrule
    baseline & The outpatient was admitted to the hospital in "postoperative of rectal cancer". & cancer\textless{}p\textgreater{}yes\textless{}p\textgreater{}postoperation\textless{}p\textgreater{}rectum                                                  \\
    ours     & The outpatient was admitted to the hospital in "postoperative of rectal cancer". & \textless{}ent\textgreater{}cancer\textless{}tendency\textgreater{}yes\textless{}character\textgreater{}postoperation\textless{}anatomy\textgreater{}rectum \\
    \bottomrule
    \end{tabular}
   \label{tab:model_example}
  \end{adjustbox}%
 \end{table}

In our model, we used special tokens corresponding to the event attribute, and formatted the output as $T=t_1A_1t_2A_2t_3A_3t_4A_4$, where $t_1$ to $t_4$ is a custom special token. In this task, as shown in Table \ref{tab:model_example}, these special tokens were represented as: \emph{<ent>}, \emph{<tency>}, \emph{<character>}, \emph{<anatomy>}. $A_1$ to $A_4$ were corresponding to "\emph{core name}", "\emph{tendency}", "\emph{characteristic}" and "\emph{anotomy}". In each events, $A_3$ and $A-4$ might be multiple options, because there were multiple entities in "\emph{characteristic}" and "\emph{anotomy}". Therefore, \emph{<unk>} tag is used as the separator. For non-existent attributes, we defined that \emph{<null>} as the null tag. We applied a model with non-special tokens as baseline model.

The experiment was fine-tuned using Mengzi-T5-base pre-training model \cite{zhang_mengzi_2021}. The dictionary size was 32128, the number of attention heads was 12, the training learning rate was 2e-5, epoch was 50, batch size was 16, the maximum input length was 256, the maximum output length was 128, and beam search length was 3. The model was stored on the validation set with lowest loss.

The indicators for this task calculate P (Precision), R (Recall), and F1 values. Multiple attributes may appear for one event per text. The event attributes is used to calculate the metrics, and all attributes need to be completely correct to calculate the F1 value. The following formula is how F1 is calculated:
\begin{equation}
      Precision=\frac{TP}{(TP+FP)}\\
      \label{eq:m_p}
  \end{equation}
  
  \begin{equation}
      Recall=\frac{TP}{(TP+FN)}\\
      \label{eq:m_r}
  \end{equation}
  
  \begin{equation}
      F_1=\frac{2 \times Precision \times Recall}{(Precision+Recall)}\\
      \label{eq:m_f1}
  \end{equation}

\section{Results}

\begin{figure}[H]
  \centering
  \includegraphics[width=7cm]{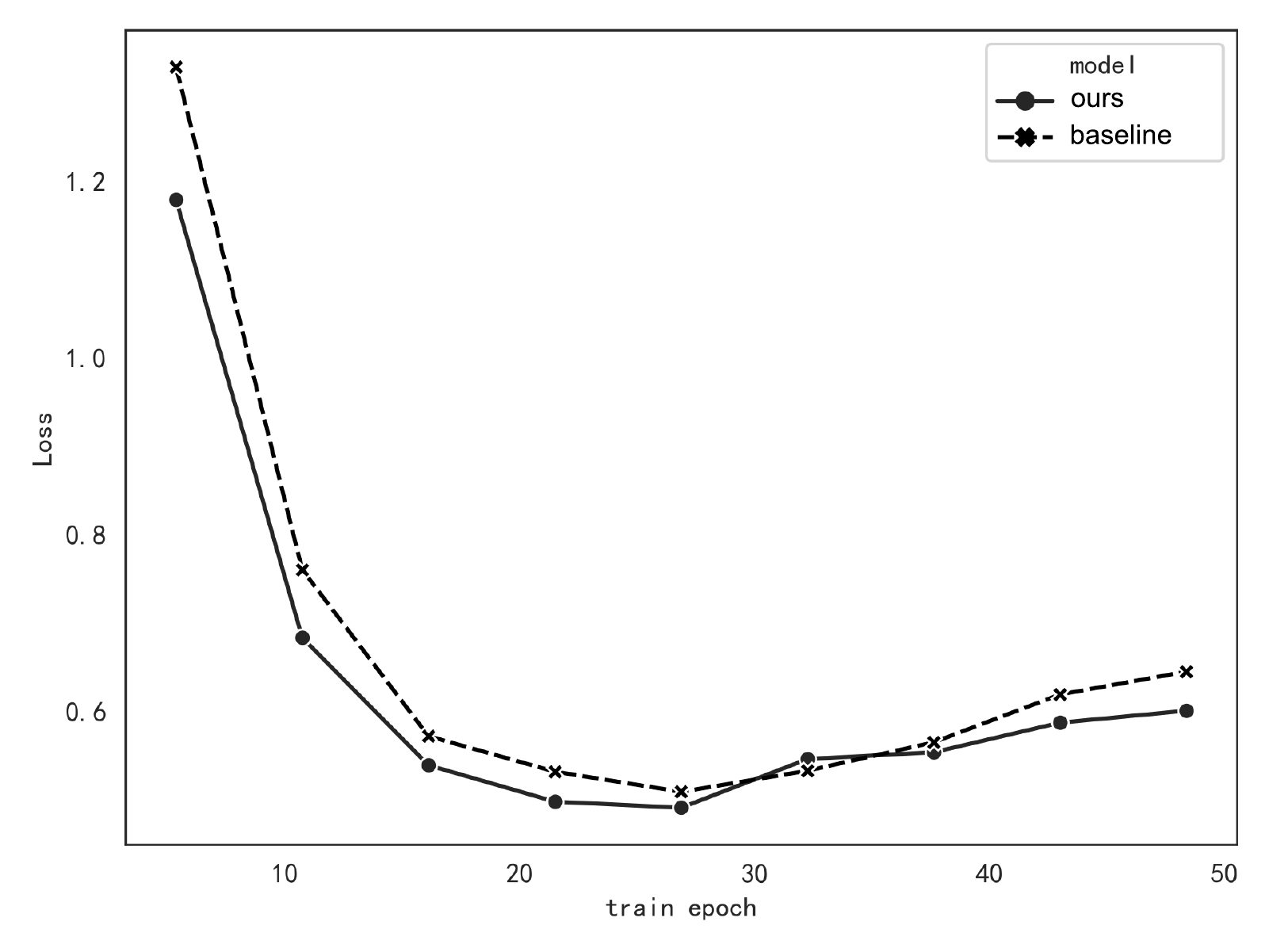}
  \caption{Training loss after each epoch.}\label{fig:loss}
\end{figure}

Figure \ref{fig:loss} shows the loss results of the models after each epoch training, and the models has converged well at the 25-th epoch. Subsequent increased in loss may be related to learning rates. Moreover, our model with type-specific characters has lower loss than baseline, indicating that type-specific characters have the ability to improve convergence of model.

\begin{table}[H]
  \caption{The proportion of events not in the original sentence.}
   \centering
   \begin{tabular}{lccccccccc}
    \toprule \\
    \multirow{2}{*}{models} & \multicolumn{3}{c}{Core words} & \multicolumn{3}{c}{Other attributes} & \multicolumn{3}{c}{Events} \\ 
    \cmidrule(rl){2-4} \cmidrule(rl){5-7} \cmidrule(rl){8-10} 
                            & P        & R        & F1       & P          & R          & F1         & P       & R       & F1     \\ \midrule
    baseline                & 0.8342   & 0.7681   & 0.7997   & 0.6342     & 0.5939     & 0.6081     & 0.5127  & 0.4720   & 0.4915 \\
    our                     & 0.8505   & 0.8047   & 0.827    & 0.6623     & 0.6266     & 0.6440      & 0.5398  & 0.5107  & 0.5248 \\ \bottomrule
    \end{tabular}
   \label{tab:performance}
 \end{table}

As shown in Table \ref{tab:performance}, our model with special tokens has best accuracy, recall, and F1 scores than the baseline model, which used the same tag to enhance characters. F1 score improved 3\% compared to the baseline model. Special tokens which represented attributes in our model had the ability to enhance events recognition. Recall score does not perform well in both models. The recall score in baseline is only 0.47, while our model is only 0.51. This indicated that there were a large number of False Negatives in the generated model and a large number of unrelated words were extracted.

% In Table \ref{tab:performance}, the results of core words and other attributes (tendencies, characteristics, and anatomies) of medical events showed that our model has relatively good output results. 
% In order to explore the reasons for the huge differences between the effect of the model in the event extraction and the core word extraction, we applied recognition criteria which required the predicted core words to be in the same order in the evaluation results (Table \ref{tab:performance_core_words}). The F1 scores dropped 0.23 in our model compared to core words metrics without strict position. 

% We can see that the model marked with special tokens is better in accuracy, recall, and F1 values than the baseline model without special tokens. Compared with the baseline model, the F1 score of our model is increased by 3\%. It indicates that type characters have the ability to enhance event recognition, so that different special tokens marks have category information. One of the main indicators of the model, the recall rate, performs poorly in both models. The recall rate in the model base model is only 0.472, while that in our model is only 0.51, which indicates that the number of false negative generated in the model is large and a large number of irrelevant words are extracted. 

\section{Error analysis}
In order to analyze the causes of errors, we verified the recognition results of core words at the same position. As shown in Table \ref{tab:performance_core_words}, although the F1 score of core words at the same position is 3.24\% higher than that of the baseline model, it is nearly 25\% lower than that of core words extracted at non fixed positions. It shows that the main reason affecting the accuracy of event extraction is that some core words are missing. The reasons for the errors are analyzed below. After checking the prediction results, it is found that the main errors are as follows.

\begin{table}[H]
  \caption{Result of models on core words with strict position.}
   \centering
   \begin{tabular}{lccc}
    \toprule
    models   & P      & R      & F1     \\ \midrule
    baseline & 0.5845 & 0.5381 & 0.5604 \\
    our      & 0.6097 & 0.5768 & 0.5928 \\ \bottomrule
    \end{tabular}
   \label{tab:performance_core_words}
 \end{table}

 Firstly, rare core words, such as "\emph{... coronary cta:1. right dominant coronary artery 2. left dominant coronary artery...}", should be extracted from the core word "\emph{right dominant coronary artery}". However, these words occur less often, so they appear with incorrect markers or missed labels.

 Secondly, core words with similar contextual structures, such as "\emph{... intermittent white phlegm, intermittent coughing of dark red blood}", should be extracted from the following events:a. The core word "\emph{cough up phlegm}", characterized by "\emph{intermittent, white, sticky}"; B. Core word "\emph{cough up blood}", characterized by "\emph{dark red, intermittent}". However, there are omissions in this model. The extracting events are: the core word "\emph{cough up phlegm}", the feature "\emph{dark red, white, sticky, intermittent}". Because the contextual structure of the two event core words is similar, the model incorrectly extracts "\emph{dark red}" as the feature of the core word "\emph{cough-phlegm}". As a result, the two events are merged into one event, which affects the extraction of the two events.

 Thirdly, the influence of the pre-training model, such as "\emph{no white pottery stool}", should be extracted as "\emph{core word: stool, tendency: negative, characteristic: white pottery stool"}, but this model is extracted as "\emph{core word: stool characteristics}". Perhaps in the pre-training model, "stool characteristics" is a common word, so when predicting the next word of the word "stool", the "characteristics" is more likely to occur than the special token \emph{<trend>}. Because "\emph{stool}" is a common core word in this dataset, this reason has a greater impact on performance.

 From the error analysis above, we can see that there are three main reasons for the omission of core words. Possible solutions include: 1. For rare core words, most of them lack professional vocabulary. By adding medical knowledge, it is expected to further increase the probability of specific vocabulary generation. 2. For the situation where the context structure of core words is close to each other, consider the extraction of separated core words and their attributes, and use the extraction of the previous step as the hint of the latter step to improve the difference between them. 3. For the effect of the pre-training model, consider constraining the generation process and strengthening the decoding process of extracting content from the original text, but may result in the loss of custom content, consider adding standardization steps.

 \section{Conclusion}
 In order to extract events from clinical records and their four attributes, in this task, we used a model generation combined with enhanced structured output to enhance event attributes with special tokens. Because our method is simple and does not require text labeling, it has great potential.

\bibliographystyle{unsrt}  
\bibliography{references}  %%% Remove comment to use the external .bib file (using bibtex).

\

\end{document}